\documentclass[final]{cvpr}

\usepackage{times}
\usepackage{epsfig}
\usepackage{graphicx}
\usepackage{amsmath}
\usepackage{amssymb}
\usepackage{subcaption}
\usepackage{booktabs}

\usepackage{algorithm}
\usepackage{algorithmic}
\usepackage{multirow}
\usepackage{tabularx}
\usepackage{verbatim}
\usepackage[pagebackref=true,breaklinks=true,colorlinks,bookmarks=false]{hyperref}

\begin{document}

\title{ConvMLP:  Hierarchical Convolutional MLPs for Vision}

\author{ Jiachen Li\textsuperscript{1,2}, Ali Hassani\textsuperscript{1}, Steven Walton\textsuperscript{1}, Humphrey Shi\textsuperscript{1,2,3} \\
{\small \textsuperscript{1}SHI Lab $@$ University of Oregon, \textsuperscript{2}UIUC, \textsuperscript{3}Picsart AI Research (PAIR)}\\
}

\maketitle

\begin{abstract}
MLP-based architectures, which consist of a sequence of consecutive multi-layer perceptron blocks, have recently been found to reach comparable results to convolutional and transformer-based methods. However, most adopt spatial MLPs which take fixed dimension inputs, therefore making it difficult to apply them to downstream tasks, such as object detection and semantic segmentation. Moreover, single-stage designs further limit performance in other computer vision tasks and fully connected layers bear heavy computation. To tackle these problems, we propose ConvMLP: a hierarchical Convolutional MLP for visual recognition, which is a light-weight, stage-wise, co-design of convolution layers, and MLPs. In particular, ConvMLP-S achieves 76.8\% top-1 accuracy on ImageNet-1k with 9M parameters and 2.4 GMACs (15\% and 19\% of MLP-Mixer-B/16, respectively).
Experiments on object detection and semantic segmentation further show that visual representation learned by ConvMLP can be seamlessly transferred and achieve competitive results with fewer parameters. 
Our code and pre-trained models are publicly available at \href{https://github.com/SHI-Labs/Convolutional-MLPs}{https://github.com/SHI-Labs/Convolutional-MLPs}
\end{abstract}

\section{Introduction}
Image classification is a fundamental problem in computer vision, and most milestone solutions in the past five years have been dominated by deep convolutional neural networks. Since late 2020 and the rise of Vision Transformer~\cite{dosovitskiy2020image}, researchers have not only been applying Transformers~\cite{vaswani2017attention} to image classification, but explored more meta-models other than convolutional neural networks. MLP-Mixer~\cite{tolstikhin2021mlp} proposes token-mixing and channel-mixing MLPs to allow communication between spatial locations and channels. ResMLP~\cite{touvron2021resmlp} uses cross-patch and cross-channel sublayers as the building block, following design of ViT. gMLP~\cite{liu2021pay} connects channel MLPs by adding spatial gating units. In essence, MLP-based models show that simple feed-forward neural networks can compete with operators like convolution and attention on image classification.

\begin{figure}[htb]
\centering
\includegraphics[width=0.5\textwidth]{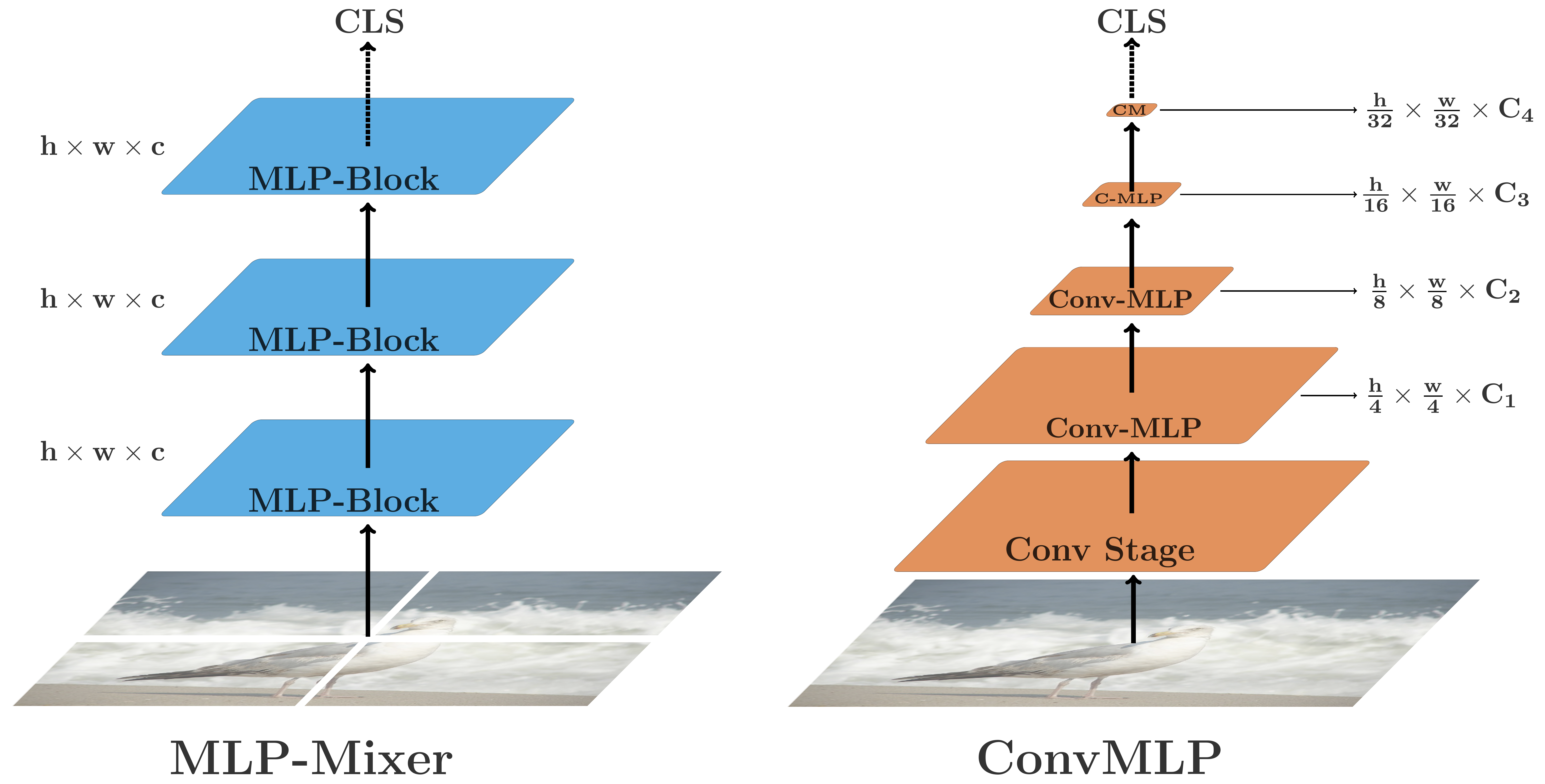}
\caption{Comparing MLP-Mixer to ConvMLP. ConvMLP adopts a simple hierarchical multi-stage co-design of convolutions and MLPs and achieves both more suitable representations as well as better accuracy vs computation trade-offs for visual recognition tasks including classification, detection and segmentation.}
\label{fig:mixerconvmlp}
\end{figure}

\begin{figure*}[htb]
\centering
\includegraphics[width=\textwidth]{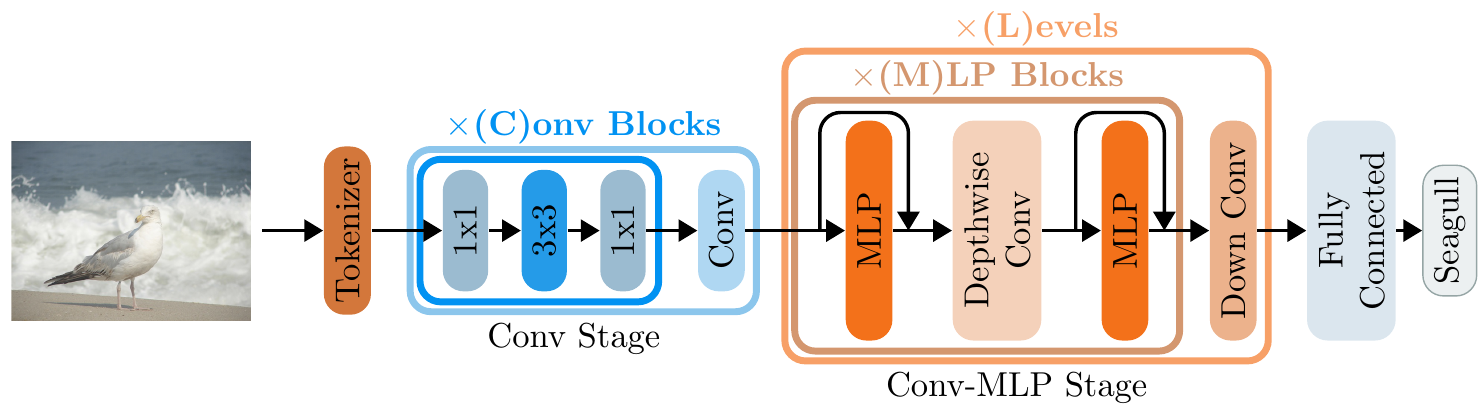}
\caption{Overview of ConvMLP framework. The Conv Stage consists of $C$ convolutional blocks with $1\times1$ and $3\times3$ kernel sizes. The MLP-Conv Stage consists of Channelwise MLPs, with skip layers, and a depthwise convolution. This is repeated $M$ times before a down convolution is utilized to express a level $L$. A level down samples an image $\mathcal{L}:h\times w\times c \mapsto \frac{h}{2^L}\times\frac{w}{2^L}\times2^L c$}
\label{fig:overview}
\end{figure*}

However, using MLPs to encode spatial information requires fixing dimension of inputs, which makes it difficult to be deployed on downstream computer vision tasks -- such as object detection and semantic segmentation -- since they usually require arbitrary resolutions of input sizes. 
Furthermore, single-stage design, following ViT, may constrain performances on object detection and semantic segmentation since they make predictions based on feature pyramids. 
Large consecutive MLPs also bring heavy computation burden and more parameters, with high dimension of hidden layers. MLP-Mixer was only able to slightly surpass ViT-Base with its large variant, which is over twice as large and twice as expensive in terms of computation. Similarly, ResMLP suffers from over 30\% more parameters and complexity, compared to a transformer-based model of similar performance.

Based on these observations, we propose ConvMLP: A Hierarchical Convolutional MLP backbone for visual recognition, which is a combination of convolution layers and MLP layers for image classification and can be seamlessly used for downstream tasks like object detection and segmentation as shown in Figure~\ref{fig:mixerconvmlp}. To remove constraints on input dimension in other MLP-like frameworks, we first replace all spatial MLPs with channel MLPs for cross-channel connections and builds a pure-MLP baseline model. To make up spatial information interaction, we add a light-weight convolution stage on top of the rest MLP stages and use convolution layers for down-sampling. Furthermore, to augment spatial connections in MLP stages, we add a simple $3\times3$ depth-wise convolution between the two channel MLPs in each MLP block, hence calling it a Conv-MLP block. This co-design of convolution layers and MLP layers builds the prototype of ConvMLP model for image classification. To make ConvMLP scalable, we extend ConvMLP model by scaling both the depth and width of both convolution and Conv-MLP stages. It achieves competitive performances on ImageNet-1k with fewer parameters compared to recent MLP-based models. On object detection and semantic segmentation, we conduct experiments on MS COCO and ADE20K benchmarks. It shows that using ConvMLP as a backbone achieves better trade-off between performance and model size.

In conclusion, our contributions are as follows:
\begin{itemize}
\item We analyze the constraints of current MLP-based models for image classification, which only take inputs of fixed dimensions and are difficult to be used in downstream computer vision tasks as backbones. Single-stage design and large computation burden further limit their applications.

\item We propose ConvMLP: a Hierachical Convolutional MLP backbone for visual recognition with co-design of convolution and MLP layers. It is scalable and can be seamlessly deployed on downstream tasks like object detection and semantic segmentation. 

\item We conduct extensive experiments on ImageNet-1k for image classification, Cifar and Flowers-102 for transfer learning, MS COCO for object detection and ADE20K for semantic segmentation to evaluate the effectiveness of our ConvMLP model.

\end{itemize}

\begin{table*}[!ht]
\centering
\renewcommand{\arraystretch}{1.6}
\begin{tabular}{c|c|c|c}
\toprule[2pt]
Stage     &ConvMLP-S &ConvMLP-M &ConvMLP-L \\ 
\midrule[1.5pt]
Conv
& \Bigg[\begin{tabular}{c}
     1$\times$1 Conv  \\
     3$\times$3 Conv  \\
     1$\times$1 Conv  \\
\end{tabular}\Bigg] $\times$ 2 
&\Bigg[\begin{tabular}{c}
     1$\times$1 Conv  \\
     3$\times$3 Conv  \\
     1$\times$1 Conv  \\
\end{tabular}\Bigg] $\times$ 3 
&\Bigg[\begin{tabular}{c}
     1$\times$1 Conv  \\
     3$\times$3 Conv  \\
     1$\times$1 Conv  \\
\end{tabular}\Bigg] $\times$ 3 \\\hline

Scale &$C_1 = 64$ &$C_1 = 64$ &$C_1 = 96$ \\ \hline
 
Conv-MLP
&\Bigg[\begin{tabular}{c}
     Channel MLP  \\
     3$\times$3 DW Conv  \\
     Channel MLP  \\
\end{tabular}\Bigg] $\times$ 2  &\Bigg[\begin{tabular}{c}
     Channel MLP  \\
     3$\times$3 DW Conv  \\
     Channel MLP  \\
\end{tabular}\Bigg] $\times$ 3 &\Bigg[\begin{tabular}{c}
     Channel MLP  \\
     3$\times$3 DW Conv  \\
     Channel MLP  \\
\end{tabular}\Bigg] $\times$ 4 \\\hline

Scale &$C_2 = 128, R=2$ &$C_2 = 128, R=3$ &$C_2 = 192, R=3$ \\ \hline

Conv-MLP
&\Bigg[\begin{tabular}{c}
     Channel MLP  \\
     3$\times$3 DW Conv  \\
     Channel MLP  \\
\end{tabular}\Bigg]   $\times$ 4 &\Bigg[\begin{tabular}{c}
     Channel MLP  \\
     3$\times$3 DW Conv  \\
     Channel MLP  \\
\end{tabular}\Bigg] $\times$ 6 &\Bigg[\begin{tabular}{c}
     Channel MLP  \\
     3$\times$3 DW Conv  \\
     Channel MLP  \\
\end{tabular}\Bigg]  $\times$ 8 \\\hline

Scale &$C_3 = 256, R=2$ &$C_3 = 256, R=3$ &$C_3 = 384, R=3$ \\ \hline

Conv-MLP
&\Bigg[\begin{tabular}{c}
     Channel MLP  \\
     3$\times$3 DW Conv  \\
     Channel MLP  \\
\end{tabular}\Bigg] $\times$ 2 &\Bigg[\begin{tabular}{c}
     Channel MLP  \\
     3$\times$3 DW Conv  \\
     Channel MLP  \\
\end{tabular}\Bigg] $\times$ 3 &\Bigg[\begin{tabular}{c}
     Channel MLP  \\
     3$\times$3 DW Conv  \\
     Channel MLP  \\
\end{tabular}\Bigg] $\times$ 3 \\\hline

Scale &$C_4 = 512, R=2$ &$C_4 = 512, R=3$ &$C_4 = 768, R=3$ \\
\bottomrule[2pt]
\end{tabular}
\caption{Detailed model architecture of ConvMLP in different scales. $R$ denotes scaling ratio of hidden layers in MLP.}
\label{tab:variants}
\end{table*}

\section{Related Work}
\noindent \textbf{Convolutional Methods} Image classification has been dominated by convolutional neural networks for almost a decade, since the rise of AlexNet~\cite{krizhevsky2012imagenet}, which introduced a convolutional neural network for image classification, and won the 2012 ILSRVC. Following that, VGGNet~\cite{simonyan2014very} proposed larger and deeper network for better performance. ResNet~\cite{he2016deep} introduced skip connections to allow training even deeper networks, and DenseNet~\cite{huang2017densely} proposed densely connected convolution layers. In the meantime, researchers explored smaller and more light-weight models that would be deployable to mobile devices. MobileNet~\cite{howard2017mobilenets,sandler2018mobilenetv2} consisted of depth-wise and point-wise convolutions, which reduced the number of parameters and computation required. ShuffleNet~\cite{ma2018shufflenet} found channel shuffling to be effective, and EfficientNet~\cite{tan2019efficientnet} further employs model scaling to width, depth and resolution for better model scalability. 

\noindent \textbf{Transformer-based Methods} Transformer~\cite{vaswani2017attention} was proposed for machine translation and has been widely adopted in most natural language processing. Recently, researchers in computer vision area adopt transformer to image classification. They propose ViT~\cite{dosovitskiy2020image} that reshapes image to patches for feature extraction by transformer encoder, which achieves comparable results to CNN-based models. DeiT~\cite{touvron2021training} further employs more data augmentation and makes ViT comparable to CNN-based model without ImageNet-22k or JFT-300M pretraining. DeiT also proposes an attention-based distillation method, which is used for student-teacher training, leading to even better performance. PVT~\cite{wang2021pyramid} proposes  feature pyramids for vision transformers, making them more compatible for downstream tasks. Swin Transformer~\cite{liu2021swin} uses patch-level multi-headed attention and stage-wise design, which also increase transferability to downstream tasks. Shuffle Swin Transformer~\cite{huang2021shuffle} proposes shuffle multi-headed attention to augment spatial connection between windows. CCT~\cite{hassani2021escaping} proposes a convolutional tokenizer and compact vision transformers, leading to better performance on smaller datasets training from scratch, with fewer parameters compared with ViT. TransCNN~\cite{liu2021transformer} also proposes a co-design of convolutions and multi-headed attention to learn hierarchical representations. 

\noindent \textbf{MLP-based Methods} MLP-Mixer~\cite{tolstikhin2021mlp} was recently proposed as a large scale image classifiers that was neither convolutional nor transformer-based. At its core, it consisted of basic matrix multiplications, data layout changes and scalar nonlinearities. ResMLP~\cite{touvron2021resmlp} followed a ResNet-like structure with MLP-based blocks instead of convolutional ones. Following that, gMLP~\cite{liu2021pay} proposed a Spatial Gating Unit to process spatial features. S$^2$-MLP~\cite{yu2021s} adopts shifted spatial feature maps to augment information communication. ViP~\cite{hou2021vision} employs linear projection on the height, width and channel dimension separately. All these methods have MLPs on fixed spatial dimensions which make it hard to be used in downstream tasks since the dimensions of spatial MLPs are fixed. Cycle MLP~\cite{chen2021cyclemlp} and AS-MLP~\cite{lian2021mlp} are concurrent works. The former replaces the spatial MLPs with cycle MLP layers and the latter with axial shifted MLPs, which make the model more flexible for varying inputs sizes. They reach competitive results on both image classification and other downstream tasks. Hire-MLP~\cite{guo2021hire} is another concurrent work that uses Hire-MLP blocks to learn hierarchical representations and achieves comparable result to transformer-based model on ImageNet.

\section{ConvMLP}

\begin{table*}[!ht]
\centering
\resizebox{0.8\textwidth}{!}{
\begin{tabular}{ccc|c|ccc}
\toprule[2pt]
Conv Stage & Conv Downsampling & Depth-Wise Conv & Epochs & \# Params (M) & GMACs & Top-1 Acc (\%) \\ 
\midrule[1.5pt]
-  & - & - & 100 & 7.88 & 1.47  & 63.29 \\
\checkmark  & - & -  & 100 & 7.89 & 1.59 & 66.69 \\
\checkmark  & \checkmark  &- & 100  & 8.71 & 1.65 & 69.56 \\
\checkmark   & - & \checkmark  & 100 & 7.91 & 1.59 & 73.84\\
\checkmark  & \checkmark & \checkmark & 100  & 8.73 & 1.65 & 74.04\\
\checkmark  & \checkmark & \checkmark & 300 & 8.73 & 1.65 & 76.33\\
\checkmark$^\dagger$  & \checkmark & \checkmark & 300 & 9.02 & 2.40 & 76.81 \\
\bottomrule[2pt]
\end{tabular}}

\caption{Ablation study on ImageNet-1k validation set. All experiments are based on ConvMLP-S. $\dagger$ denotes slightly modified Conv Stage with improved accuracy in the long run which is used in our final ConvMLP-S model.}
\label{tab:ablation}
\end{table*}
In this section, we first introduce overall design and framework of our ConvMLP. Then, we follow that design pattern including convolutional tokenizer, convolution stage and Conv-MLP Stage. We also explain how model scaling is applied to ConvMLP on convolution and Conv-MLP stages.

\subsection{Overall Design}
The overall framework of ConvMLP is illustrated in Figure~\ref{fig:overview}. Unlike other MLP-based models, we use a convolutional tokenizer to extract the initial feature map $F_1$ ($\frac{H}{4}\times \frac{W}{4} \times C_1$ dimensional). To reduce computation and improve spatial connections, we follow tokenization with a pure convolutional stage, producing feature map $F_2$ ($\frac{H}{8}\times \frac{W}{8} \times C_2$ dimensional). Then we place 3 Conv-MLP stages, generating 2 feature maps $F_3$ and $F_4$ ($\frac{H}{16}\times \frac{W}{16} \times C_3$ and $\frac{H}{32}\times \frac{W}{32} \times C_4$ dimensional respectively). Each Conv-MLP stage includes multiple Conv-MLP blocks and each Conv-MLP block has one channel MLP followed by a depth-wise convolutional layer, succeeded by another channel MLP. Similar to previous works, we include residual connections and Layer Normalization applied to inputs in the block. Each channel MLP consists of two fully connected layers with a GeLU activation~\cite{hendrycks2016gaussian} and dropout. We then apply global average pooling across to the output feature map, $F_4$, and send it through the classification head. When applying ConvMLP to downstream tasks, the feature maps $F_1$, $F_2$, $F_3$ and $F_4$ can be used to generate feature pyramids with no constraints on input size.

\subsection{Convolutional Tokenizer}
As stated, we replace the original patch tokenizer with a convolutional tokenizer. It includes three convolutional blocks, each consisting of a 3x3 convolution, batch normalization and ReLU activation. The tokenizer is also appended with a max pooling layer.

\subsection{Convolution Stage}
In order to augment spatial connections, we adopt a fully-convolutional first stage. It consists of multiple blocks, where each block is comprised of two 1x1 convolution layers with a 3x3 convolution in between.

\subsection{Conv-MLP Stage}
To reduce constraints on input dimension, we replace all spatial MLPs with channel MLPs. Since channel MLP only share weights across channels which lacks spatial interactions, we make up it by adding convolution layers in early stage, down-sampling and MLP blocks.

\noindent \textbf{Convolutional Downsampling} In the baseline model, we follow Swin Transformer~\cite{liu2021swin} that uses a patch merging method based on linear layers to down-sample feature maps. To augment adjacent spatial intersection, we replace patch merging with a 3x3 convolution layer under stride 2. It improves the classification accuracy while only brings a few more parameters.

\noindent \textbf{Convolution in MLP block} We further add a depth-wise convolution layer between two channel MLPs in one MLP block and name it Conv-MLP block. It is a 3x3 convolution layer with the same channel to the two channel MLPs, which is also used in recent Shuffle Swin Transformer~\cite{huang2021shuffle} to augment neighbor window connections. It makes up the deficiency of removing spatial MLPs, which improves the performance by a large margin while only brings few parameters.  

\subsection{Model Scaling}
To make ConvMLP scalable, we scale up ConvMLP on both width and depth of convolution stages and Conv-MLP stages.
We present 3 ConvMLP variants. Our smallest ConvMLP-S starts with only a two convolutional blocks, and has 2, 4 and 2 Conv-MLP blocks in the three Conv-MLP stages respectively. ConvMLP-M and ConvMLP-L start with three convolutional blocks. ConvMLP-M has 3, 6 and 3, and ConvMLP-L has 4, 8 and 3 Conv-MLP blocks in the three Conv-MLP stages. Details are also presented in Table~\ref{tab:variants}.
Experiments show that the performance of image classification and downstream tasks improves consistently with model scaling.

\begin{table*}[!ht]
\centering
\resizebox{\textwidth}{!}{
\begin{tabular}{l|cccc|cc}
\toprule[2pt]
Model & Backbone & \# Params (M) & GMACs & Top-1 (\%) & Acc/GMACs & Acc/MParams \\ 
\midrule[1.5pt]
\multicolumn{3}{l}{\textbf{Small models (5M-15M)}} \\
\midrule
ResNet18~\cite{he2016deep} & Convolution & 11.7 & 1.8 & 69.8 & 38.8 & 6.0 \\
Mobilenetv3~\cite{howard2019searching} & Convolution & 5.4 & 0.2 & 75.2 & 376.0 & 13.9 \\
EfficientNet-B0~\cite{tan2019efficientnet} & Convolution & 5.3 & 0.4 & 77.1 & 192.8 & 14.5 \\
\midrule
ResMLP-S12~\cite{touvron2021resmlp} & MLP & 15.3 & 3.0 & 76.6 & 25.5 & 5.0 \\
CycleMLP-B1~\cite{chen2021cyclemlp} & MLP & 15.2 & 2.1 & 78.9 & 37.6 & 5.2 \\
\midrule
ConvMLP-S (\textbf{ours}) & ConvMLP & 9.0 & 2.4 & 76.8 & 32.0 & 8.5 \\
\midrule

\multicolumn{3}{l}{\textbf{Medium-sized models (16M-30M)}} \\
\midrule
ResNet50~\cite{he2016deep} & Convolution & 25.6 & 4.1 & 76.1 & 18.6 & 3.0 \\
EfficientNet-B4~\cite{tan2019efficientnet} $\uparrow$380 & Convolution & 19.0 & 4.2 & 82.9 & 19.7 & 4.4 \\
\midrule
ViT-S~\cite{dosovitskiy2020image} $\dagger$ & Transformer & 22.1 & 4.6 & 79.9 & 17.4 & 3.6 \\
DeiT-S~\cite{touvron2021training} & Transformer & 22.1 & 4.6 & 81.2 & 17.7 & 3.7 \\
PVT-S~\cite{wang2021pyramid} & Transformer & 24.5 & 3.8 & 79.8 & 21.0 & 3.3 \\
CCT-14t~\cite{hassani2021escaping} & Transformer & 22.4 & 5.1 & 80.7 & 15.8 & 3.6 \\
\midrule
MLP-Mixer-S/16~\cite{tolstikhin2021mlp} & MLP & 18.5 & 3.8 & 73.8 & 19.4 & 4.0 \\
ResMLP-S24~\cite{touvron2021resmlp} & MLP & 30.0 & 6.0 & 79.4 & 13.2 & 2.6 \\
gMLP-S~\cite{liu2021pay} & MLP & 19.4 & 4.5 & 79.6 & 17.7 & 4.1 \\
AS-MLP-Ti~\cite{lian2021mlp} & MLP & 28.0 & 4.4 & 81.3 & 18.7 & 2.9 \\
ViP-Small/7~\cite{hou2021vision} & MLP & 25.1 & 6.9 & 81.5 & 11.8 & 3.2 \\
\midrule
ConvMLP-M (\textbf{ours}) & ConvMLP & 17.4 & 3.9 & 79.0 & 20.3 & 4.5 \\
\midrule

\multicolumn{3}{l}{\textbf{{Large models ($>$30M)}}} \\
\midrule
ResNet101~\cite{he2016deep} & Convolution & 44.6 & 7.8 & 78.0 & 10.0 & 1.7 \\
RegNetY-8GF~\cite{radosavovic2020designing} & Convolution & 39.2 & 8.0 & 79.0 & 9.9 & 2.0 \\
RegNetY-16GF~\cite{radosavovic2020designing} & Convolution & 83.6 & 15.9 & 80.4 & 5.1 & 1.0 \\
\midrule
ViT-B~\cite{dosovitskiy2020image} $\dagger$ & Transformer & 86.6 & 17.5 & 81.8 & 4.7 & 0.9 \\
DeiT-B~\cite{touvron2021training} & Transformer & 86.6 & 17.5 & 83.4 & 4.8 & 1.0 \\
PVT-L~\cite{wang2021pyramid} & Transformer & 61.4 & 9.8 & 81.7 & 8.3 & 1.3 \\
Swin Transformer-B~\cite{liu2021swin} & Transformer & 87.8 & 15.4 & 83.5 & 5.4 & 1.0 \\
Shuffle Swin-B~\cite{huang2021shuffle} & Transformer & 87.8 & 15.6 & 84.0 & 5.4 & 1.0 \\
\midrule
MLP-Mixer-B/16~\cite{tolstikhin2021mlp} & MLP & 59.9 & 12.6 & 76.4 & 6.1 & 1.3 \\
S$^2$-MLP-wide~\cite{yu2021s} & MLP & 71.0 & 14.0 & 80.0 & 5.7 & 1.1 \\
ResMLP-B24~\cite{touvron2021resmlp} & MLP & 115.7 & 23.0 & 81.0 & 3.5 & 0.7 \\
gMLP-B~\cite{liu2021pay} & MLP & 73.1 & 15.8 & 81.6 & 5.2 & 1.1 \\
ViP-Large/7~\cite{hou2021vision} & MLP & 87.8 & 24.4 & 83.2 & 3.4 & 0.9 \\
CycleMLP-B5~\cite{chen2021cyclemlp} & MLP & 75.7 & 12.3 & 83.2 & 6.7 & 0.9 \\
AS-MLP-B~\cite{lian2021mlp} & MLP & 88.0 & 15.2 & 83.3 & 5.4 & 1.0 \\
\midrule
ConvMLP-L (\textbf{ours}) & ConvMLP & 42.7 & 9.9 & 80.2 & 8.1 & 1.9 \\
\bottomrule[2pt]
\end{tabular}}
\caption{ImageNet-1k validation top-1 accuracy comparison between ConvMLP and state-of-the-art models. 
Comparing to other MLP-based methods, ConvMLP achieved the best Acc/GMACs and Acc/MParams in different model size ranges.
\textit{\small $\dagger$: reported from DeiT for fairer comparison; ViT-S was not proposed in the original paper. $\uparrow$ specifies image resolution, if different from 224\texttimes224.}}
\label{tab:imagenetclassif}
\end{table*}
\begin{table*}[!ht]
\centering
\resizebox{\textwidth}{!}{
\begin{tabular}{l|c|c|ccc}
\toprule[2pt]
Model & \# Params (M) & ImageNet-1k (\%) & CIFAR-10 (\%) & CIFAR-100 (\%) & Flowers-102 (\%) \\ 
\midrule[1.5pt]
ConvMLP-S & 9.0 & 76.8 & 98.0 & 87.4 & 99.5 \\
\midrule
ResMLP-S12~\cite{touvron2021resmlp} & 15.4 & 76.6 & 98.1 & 87.0 & 97.4  \\
ConvMLP-M & 17.4 & 79.0 & 98.6 & 89.1 & 99.5  \\
\midrule
ResMLP-S24~\cite{touvron2021resmlp} & 30.0 & 79.4 & 98.7 & 89.5 & 97.4  \\
ConvMLP-L & 42.7 & 80.2 & 98.6 & 88.6 & 99.5 \\
\midrule
ViT-B~\cite{dosovitskiy2020image} & 86.6 & 81.8 & 99.1 & 90.8 & 98.4 \\
DeiT-B~\cite{touvron2021training} & 86.6 & 83.4 & 99.1 & 91.3 & 98.9 \\
\bottomrule[2pt]
\end{tabular}}
\caption{Top-1 accuracy when pre-trained on ImageNet-1k and fine-tuned on CIFAR-10, CIFAR100 and Flowers-102. It reaches best performance on Flowers-102 among different model sizes.}
\label{tab:tarnsfer}
\end{table*}
\begin{figure*}[htpb!]
     \centering
     \begin{subfigure}[b]{0.245\textwidth}
         \centering
         \includegraphics[width=\textwidth]{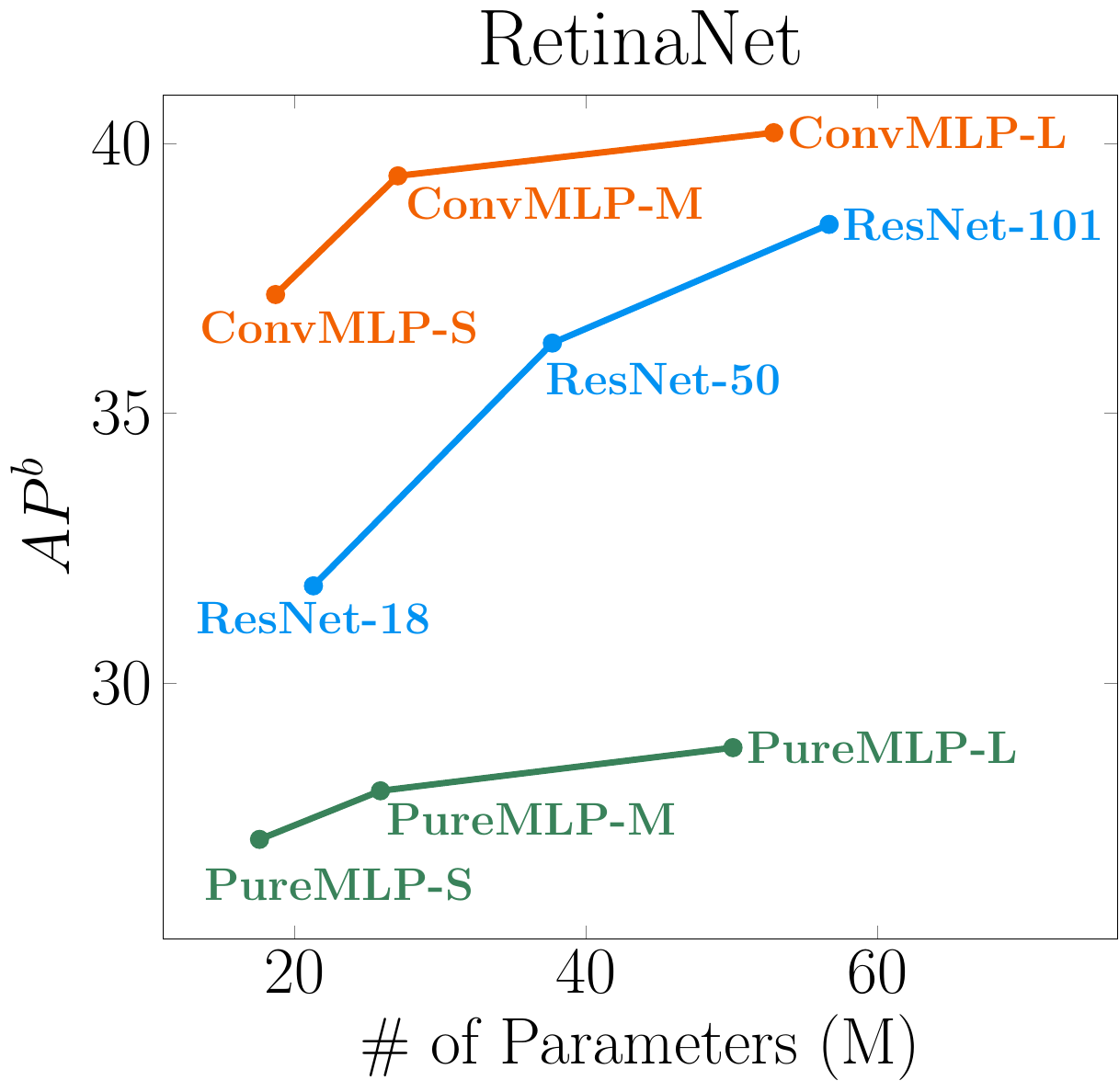}
     \end{subfigure}
     \hfill
     \begin{subfigure}[b]{0.245\textwidth}
         \centering
         \includegraphics[width=\textwidth]{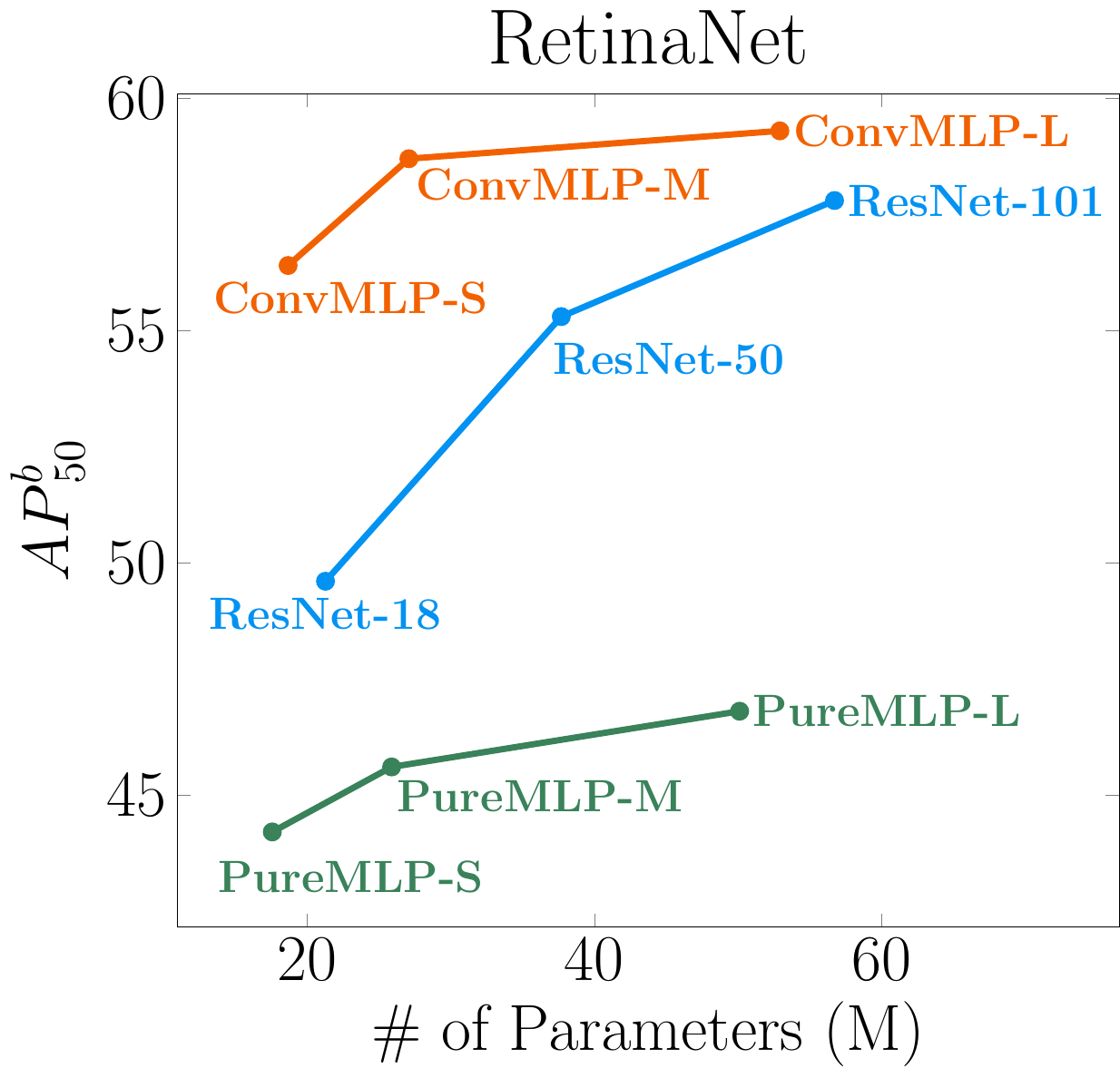}
     \end{subfigure}
     \hfill
     \begin{subfigure}[b]{0.245\textwidth}
         \centering
         \includegraphics[width=\textwidth]{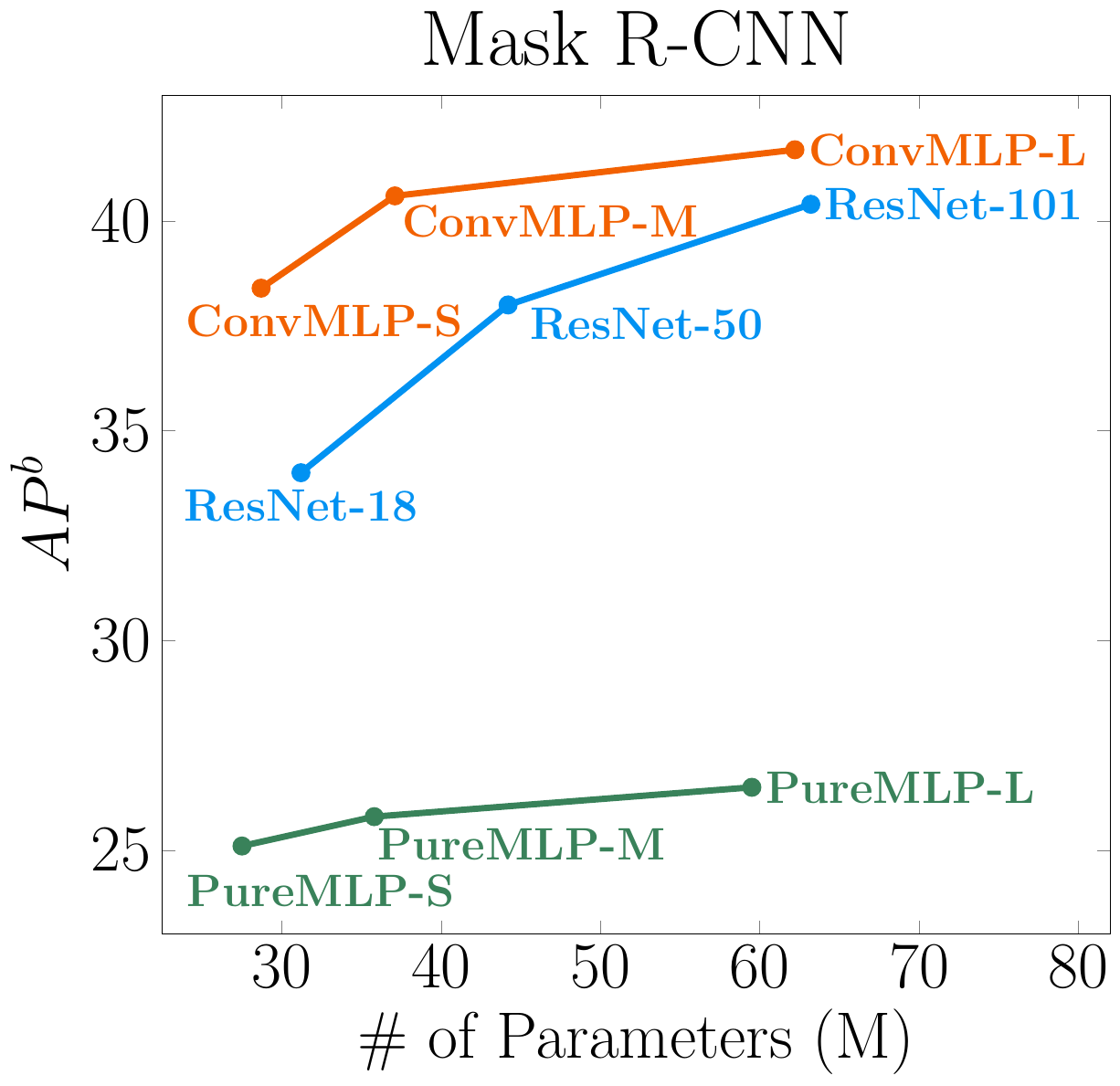}
     \end{subfigure}
     \hfill
     \begin{subfigure}[b]{0.245\textwidth}
         \centering
         \includegraphics[width=\textwidth]{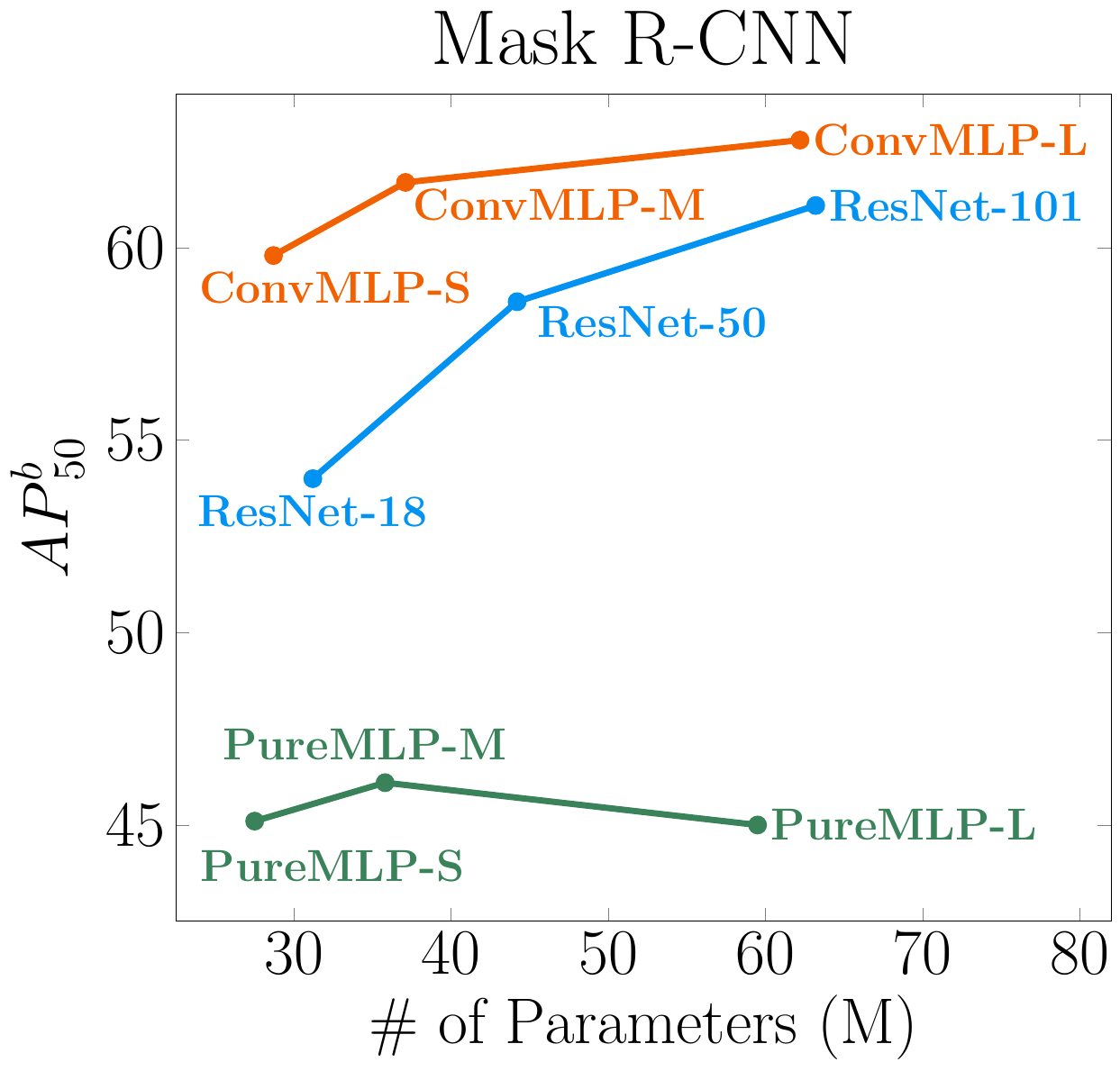}
     \end{subfigure}\\
     \begin{subfigure}[b]{0.245\textwidth}
         \centering
         \includegraphics[width=\textwidth]{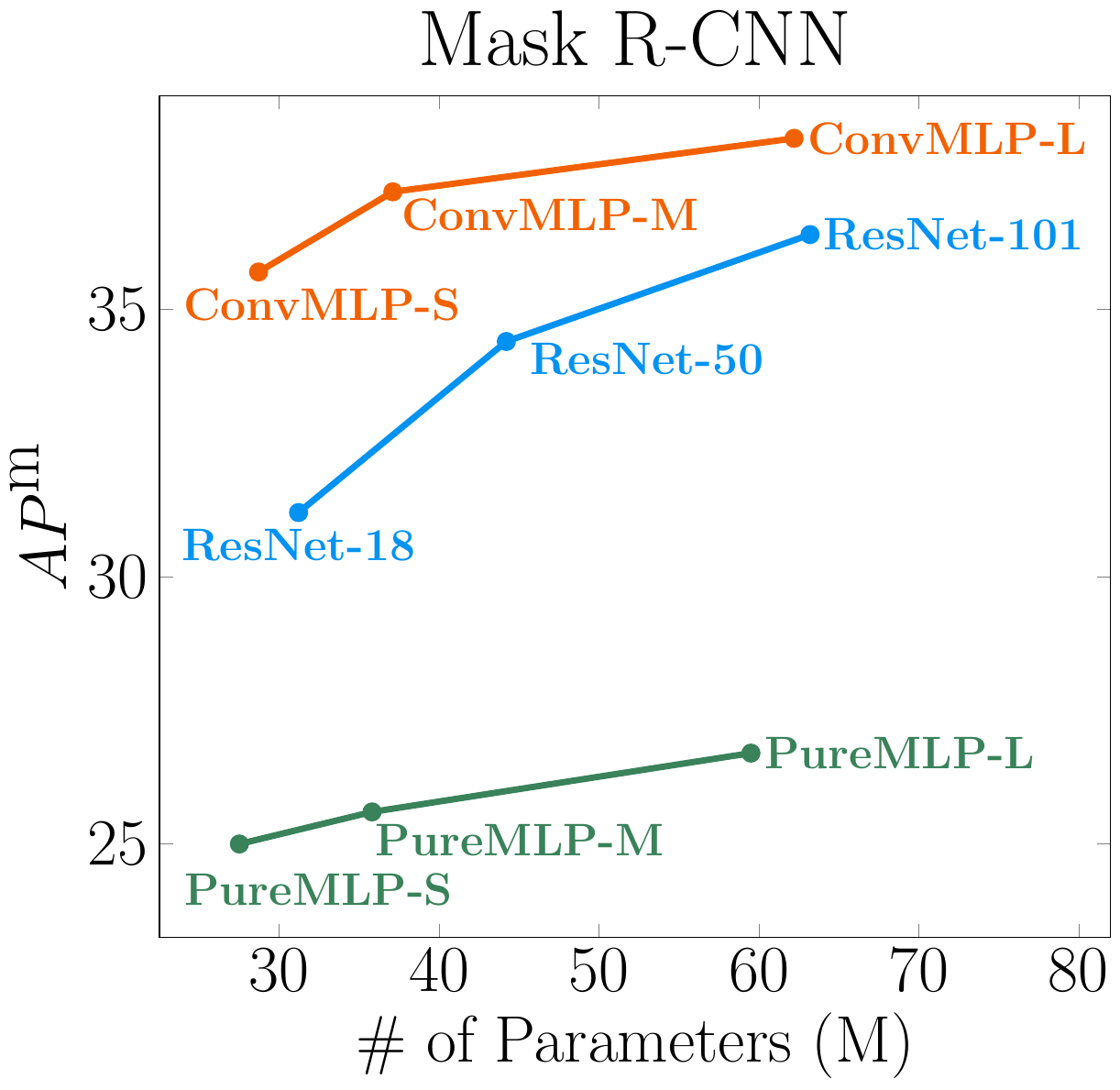}
     \end{subfigure}
     \hfill
     \begin{subfigure}[b]{0.245\textwidth}
         \centering
         \includegraphics[width=\textwidth]{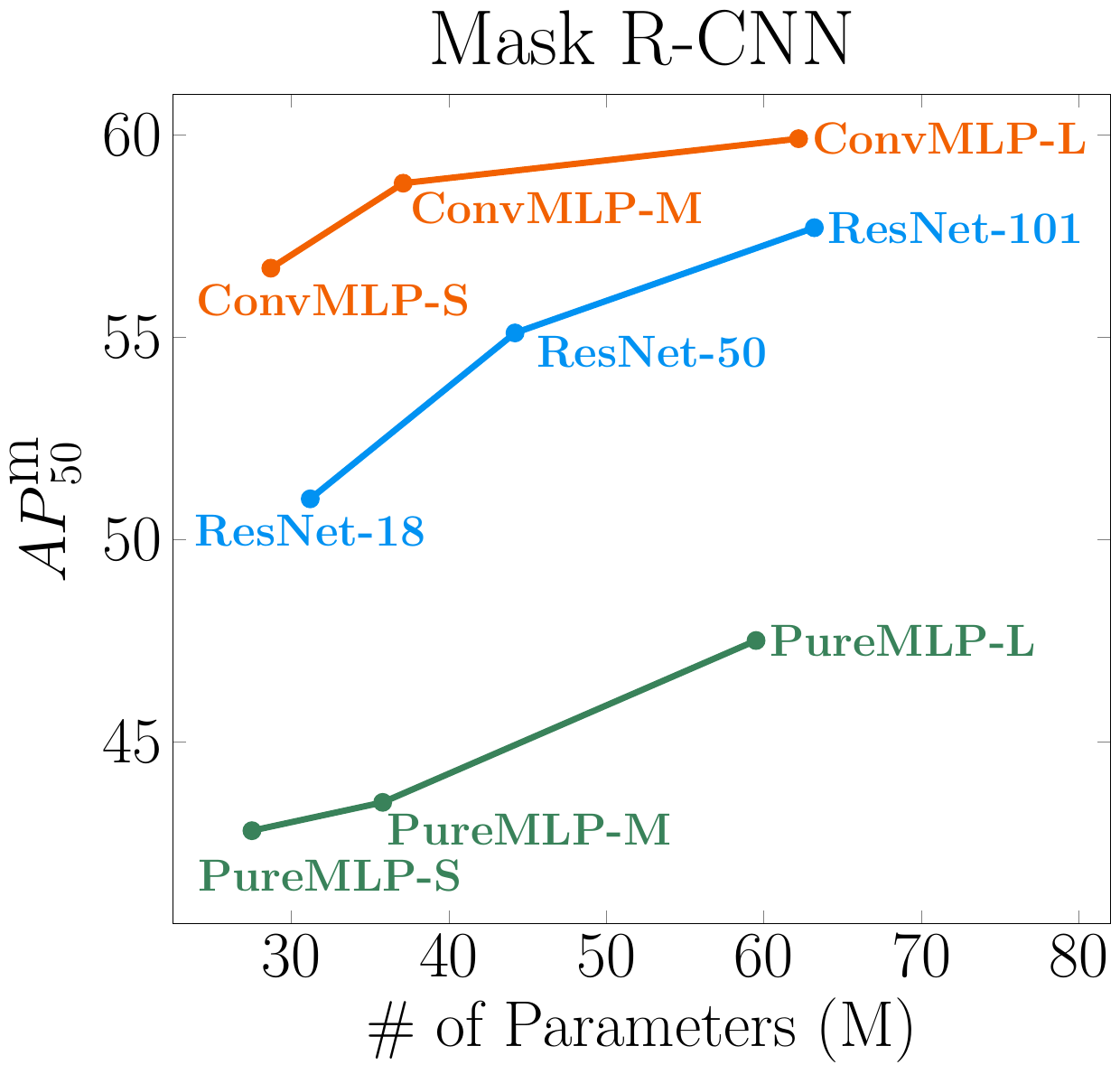}
     \end{subfigure}
     \hfill
     \begin{subfigure}[b]{0.245\textwidth}
         \centering
         \includegraphics[width=\textwidth]{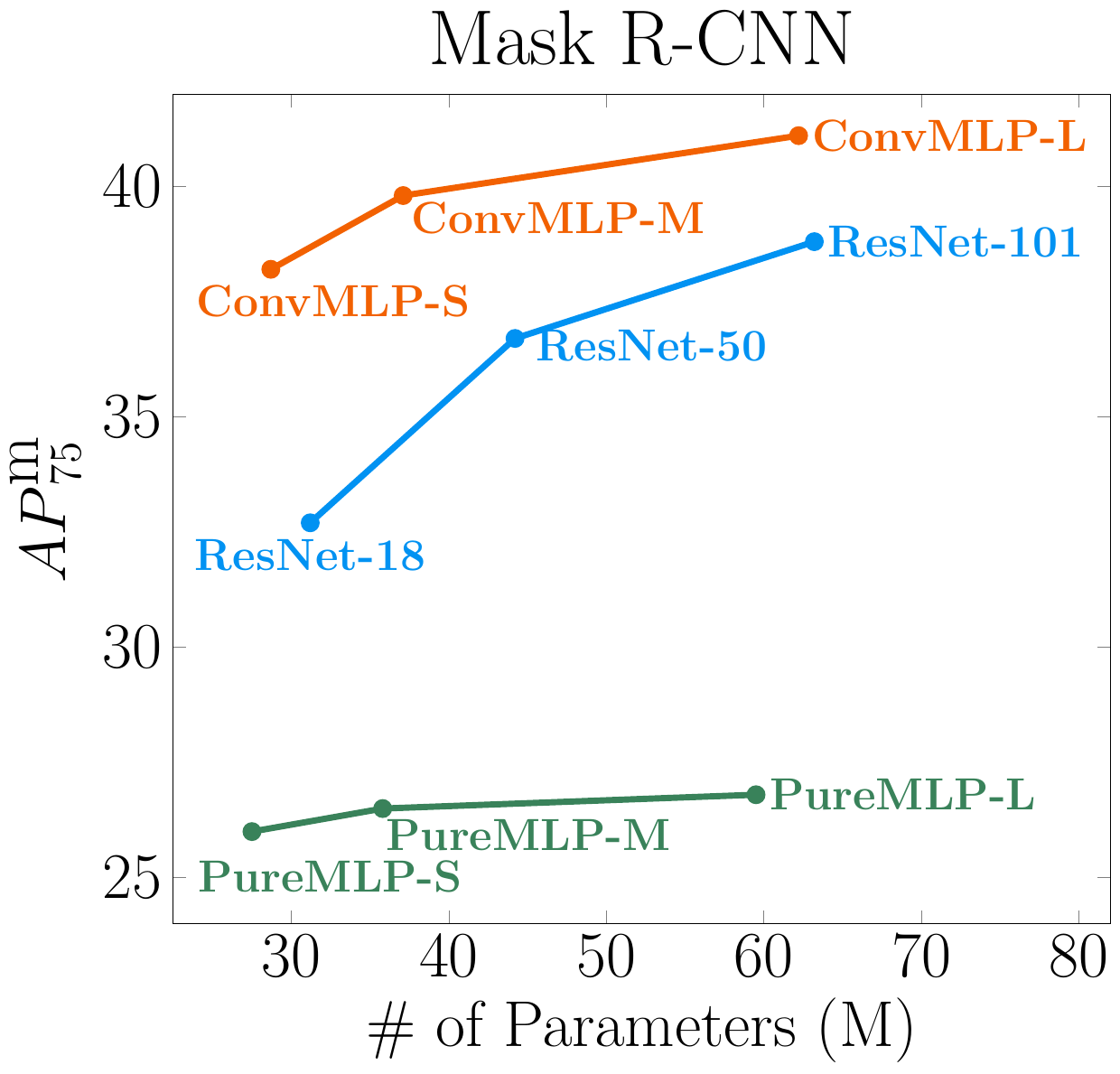}
     \end{subfigure}
     \hfill
     \begin{subfigure}[b]{0.245\textwidth}
         \centering
         \includegraphics[width=\textwidth]{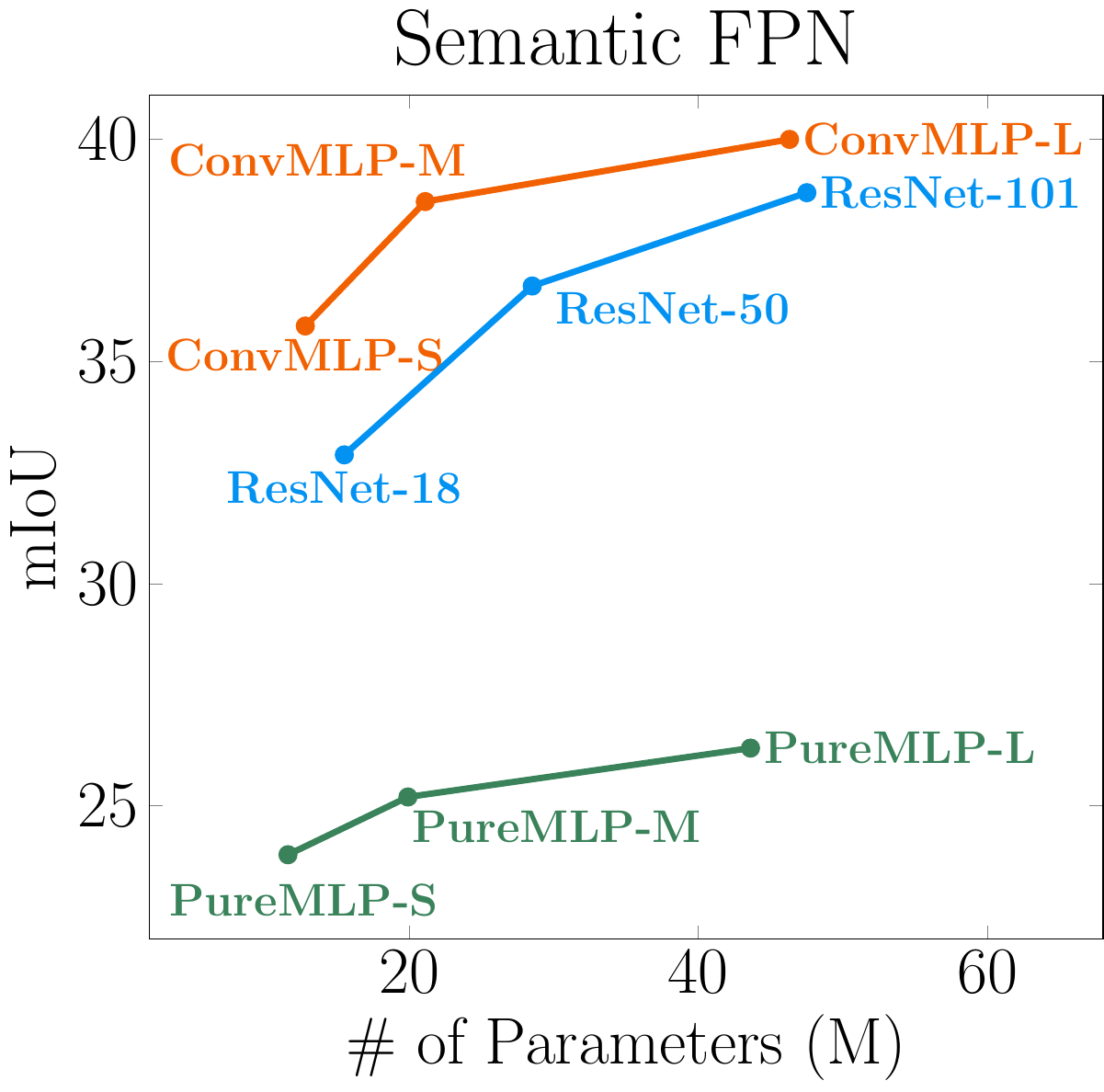}
     \end{subfigure}\\
    \caption{Comparisons between ConvMLP, Pure-MLP and ResNet as backbones of RetinaNet, Mask R-CNN on MS COCO and Semantic FPN on ADE20K. ConvMLP-based models show consistent improvements under different evaluation metrics and tasks.}
    \label{fig:downstream}
\end{figure*}

\begin{figure*}[htb]
\centering
\includegraphics[width=1\textwidth]{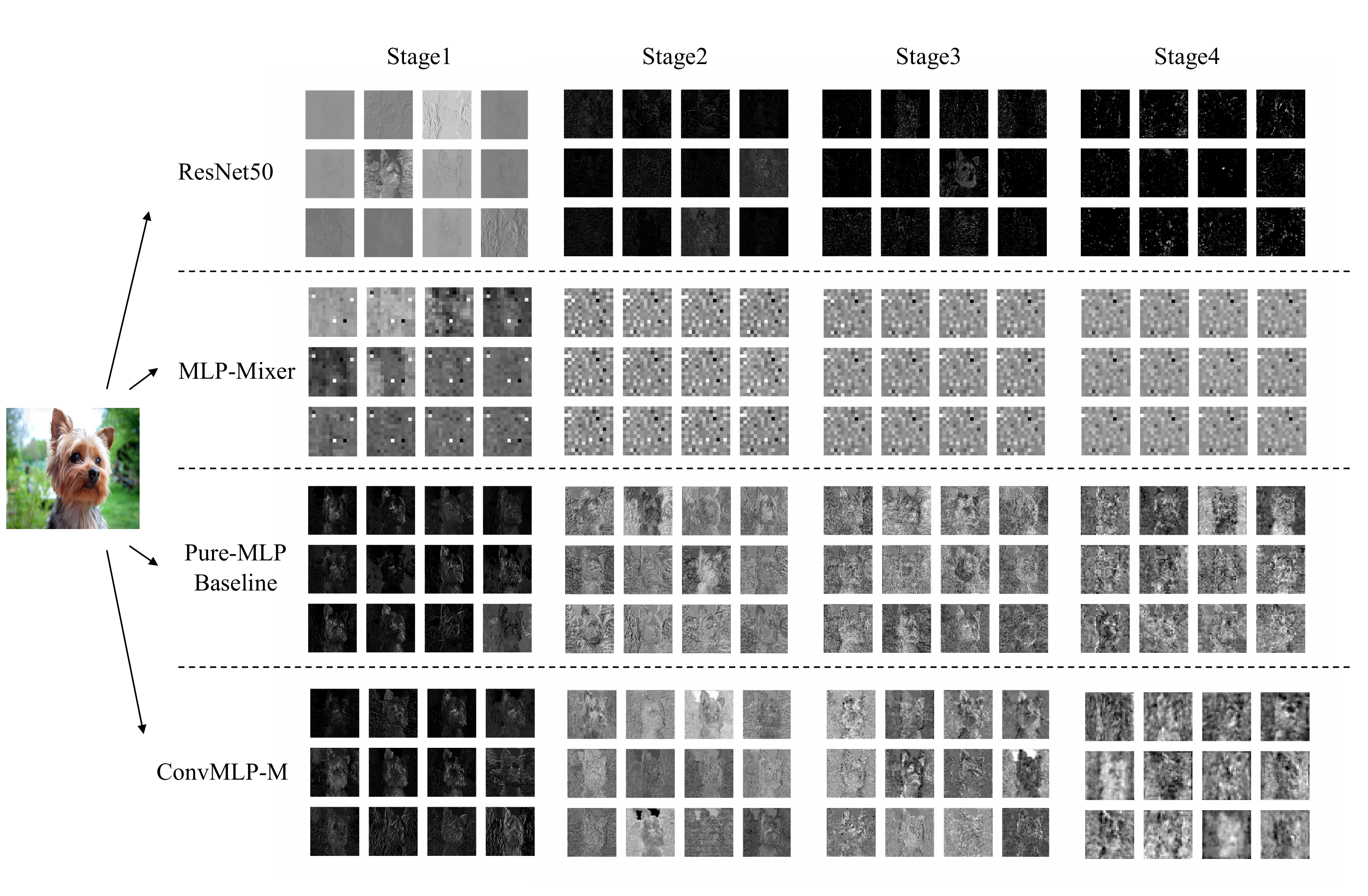}
\caption{Visualization of feature maps in different stages of ResNet50, MLP-Mixer, Pure-MLP Baseline and ConvMLP-M. Visual representations learned by ConvMLP-M show both semantic and low-level information.}
\label{fig:visualization}
\end{figure*}

\section{Experiments}
In this section, we mainly introduce our experiments on ImageNet-1k, CIFAR, Flowers-102, MS COCO and ADE20K benchmark. We first show ablation studies on different convolution modules in our ConvMLP framework to evaluate their effectiveness. Then, we compare ConvMLP to other state-of-the-art models on ImageNet-1k. We then show transferring ability on CIFAR and Flowers-102. On MS COCO and ADE20K benchmark, we use ConvMLP as backbones of RetinaNet, Mask R-CNN, Semantic FPN and it shows consistent improvements on these different downstream models.

\subsection{ImageNet-1k}
ImageNet-1k~\cite{krizhevsky2012imagenet} contains 1.2M training images and 50k images on 1000 categories for evaluating performances of classifiers. We follow standard practice provided by \verb|timm|~\cite{rw2019timm} toolbox. We use RandAugment~\cite{cubuk2020randaugment} Mixup~\cite{zhang2017mixup}, and CutMix~\cite{yun2019cutmix} for data augmentation. AdamW~\cite{loshchilov2017decoupled} is adopted as optimizer with momentum of 0.9 and weight decay of 0.05. The initial learning rate is 0.0005 with batch size of 128 on each GPU card. We use 8 NVIDIA RTX A6000 GPUs to train all models for 300 epochs and the total batch size is 1024. All other training settings and hyper-parameters are adopted from Deit~\cite{touvron2021training} for fair comparisons. For those results in ablation study, we train these models for 100 epochs with batch size 256 on each GPU and use 4 GPUs with learning rate at 0.001.

\subsection{Ablation Study}
Our baseline model Pure-MLP Baseline is composed of one patch converter and a sequence of channel MLPs in following stages. In Table~\ref{tab:ablation}, the baseline model reaches 63.29\% top-1 accuracy on ImageNet-1k and we first replace the first stage of MLPs into a convolution stage. Then, we replace the down-sampler from patch merging into a single $3\times3$ convolution layer with stride 2, which further improves top-1 accuracy to 69.56\%. To further augment spatial information communication, we add $3\times3$ depth-wise convolution between two channel MLPs and extends training epochs to 300. Finally, we modify the convolution stage with successive $1\times1$, $3\times3$, $1\times1$ convolution blocks and builds ConvMLP-S model. 

\subsection{Comparisons with SOTA}
In Table~\ref{tab:imagenetclassif}, we compare ConvMLP to other state-of-the-art image classification models on ImageNet-1k. We include Convolution-based, Transformer-based and MLP-based methods under different scales. We also present number of parameters and GMACs of these models. 

\subsection{Transfer learning}
\noindent \textbf{Dataset} We use CIFAR-10/CIFAR-100~\cite{krizhevsky2009learning} and Flowers-102~\cite{nilsback2008automated} to evaluate transferring ability of ImageNet-pretrained ConvMLP variants. Each model was fine-tuned for 50 epochs with a learning rate of 3e-4 (with cosine scheduler), weight decay of 5e-2, 10 warmup and cooldown epochs. We used the same training script and therefore augmentations as the ImageNet-1k experiments. We also resized all images to 224\texttimes224.

\noindent \textbf{Results} The results are presented in Table \ref{tab:tarnsfer}. We report results from ResMLP, ViT and DeiT as well.

\subsection{Object Detection}
\noindent \textbf{Dataset} MS COCO~\cite{lin2014microsoft} is a widely-used benchmark for evaluating object detection model. It has 118k images for training and 5k images for evaluating performances of object detectors.
We follow standard practice of RetinaNet~\cite{lin2017focal} and Mask R-CNN~\cite{he2017mask} with ResNet as backbones in \verb|mmdetection|~\cite{chen2019mmdetection}. We replace ResNet backbones with ConvMLP and adjust the dimension of convolution layers in feature pyramids accordingly. We also replace SGD optimizer with AdamW and adjust learning rate to 0.0001 with weight decay at 0.0001, which follows the configs in PVT~\cite{wang2021pyramid}. We train both RetinaNet and Mask R-CNN for 12 epochs on 8 GPUs with total batch size of 16.

\noindent \textbf{Results} We transfer ResNet, Pure-MLP and ConvMLP variants to object detection on MS COCO and the results are presented in Figure~\ref{fig:downstream}. It can be observed that ConvMLP achieves better performance on object detection and instance segmentation consistently as backbones of RetinaNet and Mask R-CNN compared with Pure-MLP and ResNet. More details of the results are presented in Appendix.

\subsection{Semantic Segmentation}
\noindent \textbf{Dataset} ADE20K~\cite{zhou2017scene} is a widely-used dataset for semantic segmentation, which has 20k images for training and 2k images for evaluating the performance of semantic segmentation models. We employ standard practice of Semantic FPN~\cite{kirillov2019panoptic} implemented based on \verb|mmsegmentation|~\cite{mmseg2020}. Following PVT in semantic segmentation, we train ConvMLP-based Semantic FPN on 8 GPUs with total batch size of 16 for 40k iterations. We also replace optimizer from SGD to AdamW with learning rate at 0.0002 and weight decay at 0.0001. The learning rate decays with polynomial rate at 0.9 and input images are randomly resized and cropped to $512 \times 512$.

\noindent \textbf{Results} All experimental results on ADE20K are presented in Figure~\ref{fig:downstream}. Similar to the results of object detection, it shows that visual representations learned by ConvMLP can be transferred to pixel-level prediction task like semantic segmentation. More details of the results can be found in Appendix.

\subsection{Visualization}
We visualize feature maps of ResNet50, MLP-Mixer-B/16, Pure-MLP Baseline and ConvMLP-M under $(1024, 1024)$ input size (MLP-Mixer-B/16 under $(224,224)$ due to dimension constraint) in Figure~\ref{fig:visualization} to analyze the differences in visual representations learned by these models, and similar feature maps of transformer-based model are presented in T2T-ViT~\cite{yuan2021tokens}. We observe that representations learned by ConvMLP involve more low-level features like edges or textures compared with ResNet and more semantics compared with Pure-MLP Baseline .

\section{Conclusion}
In this paper, we analyze the constraints of current MLP-based models for visual representation learning: 1. Spatial MLPs only take inputs with fixed resolutions, making transfer to downstream tasks, such as object detection and segmentation, difficult. 2. Single-stage design and fully connected layers further constrain usage due to the added complexity. To tackle these problems, we propose ConvMLP: a Hierarchical Convolutional MLP for visual representation learning through combining convolutional layers and MLPs. The architecture can be seamlessly prepended to downstream networks like RetinaNet, Mask R-CNN and Semantic FPN. Experiments further show that it can achieve competitive results on different benchmarks with fewer parameters compared to other methods. 
The main limitation of ConvMLP is that ImageNet performance scales slower with model size. We leave this to be explored in future works.

{\small
\bibliographystyle{ieee_fullname}
\bibliography{references}

\begin{thebibliography}{10}\itemsep=-1pt

\bibitem{chen2019mmdetection}
Kai Chen, Jiaqi Wang, Jiangmiao Pang, Yuhang Cao, Yu Xiong, Xiaoxiao Li,
  Shuyang Sun, Wansen Feng, Ziwei Liu, Jiarui Xu, et~al.
\newblock Mmdetection: Open mmlab detection toolbox and benchmark.
\newblock {\em arXiv preprint arXiv:1906.07155}, 2019.

\bibitem{chen2021cyclemlp}
Shoufa Chen, Enze Xie, Chongjian Ge, Ding Liang, and Ping Luo.
\newblock Cyclemlp: A mlp-like architecture for dense prediction.
\newblock {\em arXiv preprint arXiv:2107.10224}, 2021.

\bibitem{mmseg2020}
MMSegmentation Contributors.
\newblock {MMSegmentation}: Openmmlab semantic segmentation toolbox and
  benchmark.
\newblock \url{https://github.com/open-mmlab/mmsegmentation}, 2020.

\bibitem{cubuk2020randaugment}
Ekin~D Cubuk, Barret Zoph, Jonathon Shlens, and Quoc~V Le.
\newblock Randaugment: Practical automated data augmentation with a reduced
  search space.
\newblock In {\em Proceedings of the IEEE/CVF Conference on Computer Vision and
  Pattern Recognition Workshops}, pages 702--703, 2020.

\bibitem{dosovitskiy2020image}
Alexey Dosovitskiy, Lucas Beyer, Alexander Kolesnikov, Dirk Weissenborn,
  Xiaohua Zhai, Thomas Unterthiner, Mostafa Dehghani, Matthias Minderer, Georg
  Heigold, Sylvain Gelly, et~al.
\newblock An image is worth 16x16 words: Transformers for image recognition at
  scale.
\newblock {\em arXiv preprint arXiv:2010.11929}, 2020.

\bibitem{guo2021hire}
Jianyuan Guo, Yehui Tang, Kai Han, Xinghao Chen, Han Wu, Chao Xu, Chang Xu, and
  Yunhe Wang.
\newblock Hire-mlp: Vision mlp via hierarchical rearrangement.
\newblock {\em arXiv preprint arXiv:2108.13341}, 2021.

\bibitem{hassani2021escaping}
Ali Hassani, Steven Walton, Nikhil Shah, Abulikemu Abuduweili, Jiachen Li, and
  Humphrey Shi.
\newblock Escaping the big data paradigm with compact transformers.
\newblock {\em arXiv preprint arXiv:2104.05704}, 2021.

\bibitem{he2017mask}
Kaiming He, Georgia Gkioxari, Piotr Doll{\'a}r, and Ross Girshick.
\newblock Mask r-cnn.
\newblock In {\em Proceedings of the IEEE international conference on computer
  vision}, pages 2961--2969, 2017.

\bibitem{he2016deep}
Kaiming He, Xiangyu Zhang, Shaoqing Ren, and Jian Sun.
\newblock Deep residual learning for image recognition.
\newblock In {\em Proceedings of the IEEE conference on computer vision and
  pattern recognition}, pages 770--778, 2016.

\bibitem{hendrycks2016gaussian}
Dan Hendrycks and Kevin Gimpel.
\newblock Gaussian error linear units (gelus).
\newblock {\em arXiv preprint arXiv:1606.08415}, 2016.

\bibitem{hou2021vision}
Qibin Hou, Zihang Jiang, Li Yuan, Ming-Ming Cheng, Shuicheng Yan, and Jiashi
  Feng.
\newblock Vision permutator: A permutable mlp-like architecture for visual
  recognition.
\newblock {\em arXiv preprint arXiv:2106.12368}, 2021.

\bibitem{howard2019searching}
Andrew Howard, Mark Sandler, Grace Chu, Liang-Chieh Chen, Bo Chen, Mingxing
  Tan, Weijun Wang, Yukun Zhu, Ruoming Pang, Vijay Vasudevan, et~al.
\newblock Searching for mobilenetv3.
\newblock In {\em Proceedings of the IEEE/CVF International Conference on
  Computer Vision}, pages 1314--1324, 2019.

\bibitem{howard2017mobilenets}
Andrew~G Howard, Menglong Zhu, Bo Chen, Dmitry Kalenichenko, Weijun Wang,
  Tobias Weyand, Marco Andreetto, and Hartwig Adam.
\newblock Mobilenets: Efficient convolutional neural networks for mobile vision
  applications.
\newblock {\em arXiv preprint arXiv:1704.04861}, 2017.

\bibitem{huang2017densely}
Gao Huang, Zhuang Liu, Laurens Van Der~Maaten, and Kilian~Q Weinberger.
\newblock Densely connected convolutional networks.
\newblock In {\em Proceedings of the IEEE conference on computer vision and
  pattern recognition}, pages 4700--4708, 2017.

\bibitem{huang2021shuffle}
Zilong Huang, Youcheng Ben, Guozhong Luo, Pei Cheng, Gang Yu, and Bin Fu.
\newblock Shuffle transformer: Rethinking spatial shuffle for vision
  transformer.
\newblock {\em arXiv preprint arXiv:2106.03650}, 2021.

\bibitem{kirillov2019panoptic}
Alexander Kirillov, Ross Girshick, Kaiming He, and Piotr Doll{\'a}r.
\newblock Panoptic feature pyramid networks.
\newblock In {\em Proceedings of the IEEE/CVF Conference on Computer Vision and
  Pattern Recognition}, pages 6399--6408, 2019.

\bibitem{krizhevsky2009learning}
Alex Krizhevsky, Geoffrey Hinton, et~al.
\newblock Learning multiple layers of features from tiny images, 2009.

\bibitem{krizhevsky2012imagenet}
Alex Krizhevsky, Ilya Sutskever, and Geoffrey~E Hinton.
\newblock Imagenet classification with deep convolutional neural networks.
\newblock {\em Advances in neural information processing systems},
  25:1097--1105, 2012.

\bibitem{lian2021mlp}
Dongze Lian, Zehao Yu, Xing Sun, and Shenghua Gao.
\newblock As-mlp: An axial shifted mlp architecture for vision.
\newblock {\em arXiv preprint arXiv:2107.08391}, 2021.

\bibitem{lin2017focal}
Tsung-Yi Lin, Priya Goyal, Ross Girshick, Kaiming He, and Piotr Doll{\'a}r.
\newblock Focal loss for dense object detection.
\newblock In {\em Proceedings of the IEEE international conference on computer
  vision}, pages 2980--2988, 2017.

\bibitem{lin2014microsoft}
Tsung-Yi Lin, Michael Maire, Serge Belongie, James Hays, Pietro Perona, Deva
  Ramanan, Piotr Doll{\'a}r, and C~Lawrence Zitnick.
\newblock Microsoft coco: Common objects in context.
\newblock In {\em European conference on computer vision}, pages 740--755.
  Springer, 2014.

\bibitem{liu2021pay}
Hanxiao Liu, Zihang Dai, David~R So, and Quoc~V Le.
\newblock Pay attention to mlps.
\newblock {\em arXiv preprint arXiv:2105.08050}, 2021.

\bibitem{liu2021transformer}
Yun Liu, Guolei Sun, Yu Qiu, Le Zhang, Ajad Chhatkuli, and Luc Van~Gool.
\newblock Transformer in convolutional neural networks.
\newblock {\em arXiv preprint arXiv:2106.03180}, 2021.

\bibitem{liu2021swin}
Ze Liu, Yutong Lin, Yue Cao, Han Hu, Yixuan Wei, Zheng Zhang, Stephen Lin, and
  Baining Guo.
\newblock Swin transformer: Hierarchical vision transformer using shifted
  windows.
\newblock {\em arXiv preprint arXiv:2103.14030}, 2021.

\bibitem{loshchilov2017decoupled}
Ilya Loshchilov and Frank Hutter.
\newblock Decoupled weight decay regularization.
\newblock {\em arXiv preprint arXiv:1711.05101}, 2017.

\bibitem{ma2018shufflenet}
Ningning Ma, Xiangyu Zhang, Hai-Tao Zheng, and Jian Sun.
\newblock Shufflenet v2: Practical guidelines for efficient cnn architecture
  design.
\newblock In {\em Proceedings of the European conference on computer vision
  (ECCV)}, pages 116--131, 2018.

\bibitem{nilsback2008automated}
Maria-Elena Nilsback and Andrew Zisserman.
\newblock Automated flower classification over a large number of classes.
\newblock In {\em 2008 Sixth Indian Conference on Computer Vision, Graphics
  Image Processing}, pages 722--729, 2008.

\bibitem{radosavovic2020designing}
Ilija Radosavovic, Raj~Prateek Kosaraju, Ross Girshick, Kaiming He, and Piotr
  Doll{\'a}r.
\newblock Designing network design spaces.
\newblock In {\em Proceedings of the IEEE/CVF Conference on Computer Vision and
  Pattern Recognition}, pages 10428--10436, 2020.

\bibitem{ribeiro2016why}
Marco~T{\'{u}}lio Ribeiro, Sameer Singh, and Carlos Guestrin.
\newblock "why should {I} trust you?": Explaining the predictions of any
  classifier.
\newblock {\em CoRR}, abs/1602.04938, 2016.

\bibitem{sandler2018mobilenetv2}
Mark Sandler, Andrew Howard, Menglong Zhu, Andrey Zhmoginov, and Liang-Chieh
  Chen.
\newblock Mobilenetv2: Inverted residuals and linear bottlenecks.
\newblock In {\em Proceedings of the IEEE conference on computer vision and
  pattern recognition}, pages 4510--4520, 2018.

\bibitem{simonyan2014very}
Karen Simonyan and Andrew Zisserman.
\newblock Very deep convolutional networks for large-scale image recognition.
\newblock {\em arXiv preprint arXiv:1409.1556}, 2014.

\bibitem{tan2019efficientnet}
Mingxing Tan and Quoc Le.
\newblock Efficientnet: Rethinking model scaling for convolutional neural
  networks.
\newblock In {\em International Conference on Machine Learning}, pages
  6105--6114. PMLR, 2019.

\bibitem{tolstikhin2021mlp}
Ilya Tolstikhin, Neil Houlsby, Alexander Kolesnikov, Lucas Beyer, Xiaohua Zhai,
  Thomas Unterthiner, Jessica Yung, Daniel Keysers, Jakob Uszkoreit, Mario
  Lucic, et~al.
\newblock Mlp-mixer: An all-mlp architecture for vision.
\newblock {\em arXiv preprint arXiv:2105.01601}, 2021.

\bibitem{touvron2021resmlp}
Hugo Touvron, Piotr Bojanowski, Mathilde Caron, Matthieu Cord, Alaaeldin
  El-Nouby, Edouard Grave, Armand Joulin, Gabriel Synnaeve, Jakob Verbeek, and
  Herv{\'e} J{\'e}gou.
\newblock Resmlp: Feedforward networks for image classification with
  data-efficient training.
\newblock {\em arXiv preprint arXiv:2105.03404}, 2021.

\bibitem{touvron2021training}
Hugo Touvron, Matthieu Cord, Matthijs Douze, Francisco Massa, Alexandre
  Sablayrolles, and Herv{\'e} J{\'e}gou.
\newblock Training data-efficient image transformers \& distillation through
  attention.
\newblock In {\em International Conference on Machine Learning}, pages
  10347--10357. PMLR, 2021.

\bibitem{vaswani2017attention}
Ashish Vaswani, Noam Shazeer, Niki Parmar, Jakob Uszkoreit, Llion Jones,
  Aidan~N Gomez, {\L}ukasz Kaiser, and Illia Polosukhin.
\newblock Attention is all you need.
\newblock In {\em Advances in neural information processing systems}, pages
  5998--6008, 2017.

\bibitem{wang2021pyramid}
Wenhai Wang, Enze Xie, Xiang Li, Deng-Ping Fan, Kaitao Song, Ding Liang, Tong
  Lu, Ping Luo, and Ling Shao.
\newblock Pyramid vision transformer: A versatile backbone for dense prediction
  without convolutions.
\newblock {\em arXiv preprint arXiv:2102.12122}, 2021.

\bibitem{rw2019timm}
Ross Wightman.
\newblock Pytorch image models.
\newblock \url{https://github.com/rwightman/pytorch-image-models}, 2019.

\bibitem{yu2021s}
Tan Yu, Xu Li, Yunfeng Cai, Mingming Sun, and Ping Li.
\newblock S$^2$-mlp: Spatial-shift mlp architecture for vision.
\newblock {\em arXiv preprint arXiv:2106.07477}, 2021.

\bibitem{yuan2021tokens}
Li Yuan, Yunpeng Chen, Tao Wang, Weihao Yu, Yujun Shi, Zihang Jiang, Francis~EH
  Tay, Jiashi Feng, and Shuicheng Yan.
\newblock Tokens-to-token vit: Training vision transformers from scratch on
  imagenet.
\newblock {\em arXiv preprint arXiv:2101.11986}, 2021.

\bibitem{yun2019cutmix}
Sangdoo Yun, Dongyoon Han, Seong~Joon Oh, Sanghyuk Chun, Junsuk Choe, and
  Youngjoon Yoo.
\newblock Cutmix: Regularization strategy to train strong classifiers with
  localizable features.
\newblock In {\em Proceedings of the IEEE/CVF International Conference on
  Computer Vision}, pages 6023--6032, 2019.

\bibitem{zhang2017mixup}
Hongyi Zhang, Moustapha Cisse, Yann~N Dauphin, and David Lopez-Paz.
\newblock mixup: Beyond empirical risk minimization.
\newblock {\em arXiv preprint arXiv:1710.09412}, 2017.

\bibitem{zhou2017scene}
Bolei Zhou, Hang Zhao, Xavier Puig, Sanja Fidler, Adela Barriuso, and Antonio
  Torralba.
\newblock Scene parsing through ade20k dataset.
\newblock In {\em Proceedings of the IEEE conference on computer vision and
  pattern recognition}, pages 633--641, 2017.

\end{thebibliography}
}

\appendix

\begin{figure*}[htb]
\centering
\includegraphics[width=0.98\textwidth]{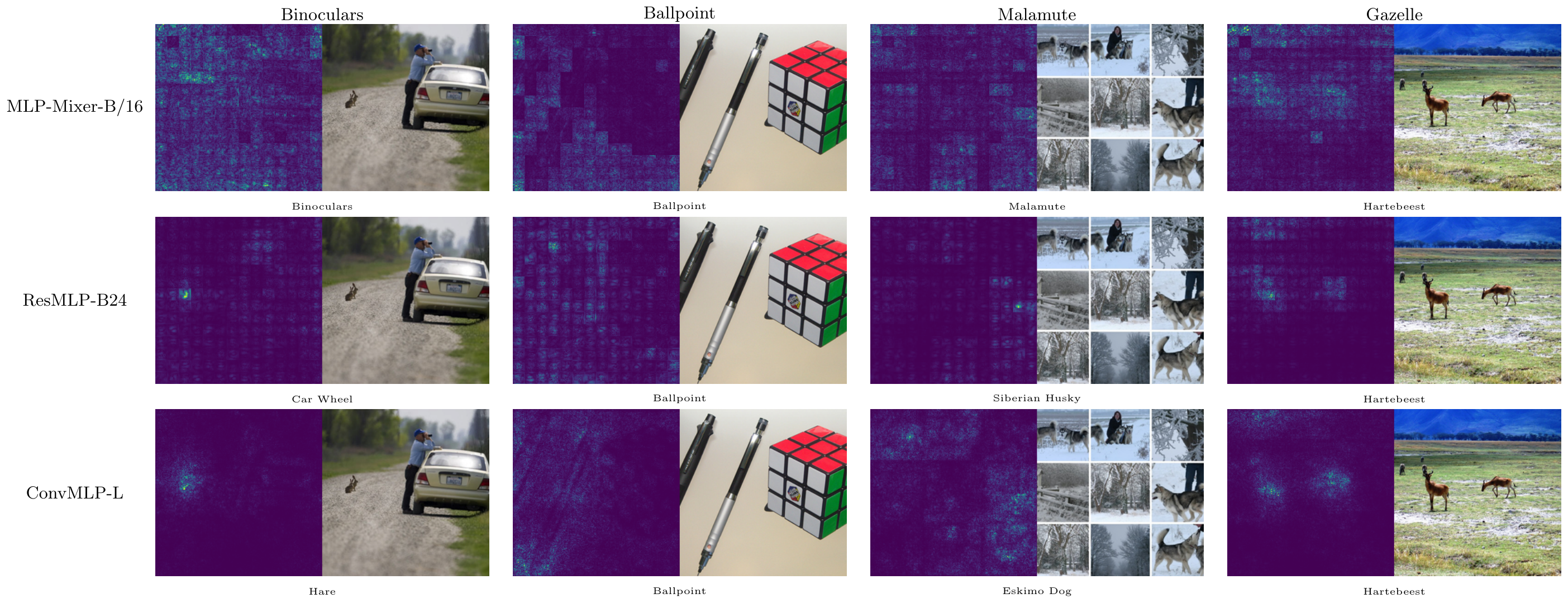}
\caption{Salient Maps of selected ImageNet images, comparing MLP-Mixer-B/16, ResMLP-B24, and ConvMLP-L. The labels at the top represent the ground truth label and the smaller labels below the images show the network's prediction.}
\label{fig:salient}
\end{figure*}

\begin{table*}[!ht]
\centering
\begin{tabular}{l|c|ccc|ccc}
\toprule[2pt]
RetinaNet Backbone & \# Params & $AP^{b}$ & $AP^{b}_{50}$ & $AP^{b}_{75}$ & $AP^{b}_{S}$ & $AP^{b}_{M}$ & $AP^{b}_{L}$ \\ 
\midrule[1.5pt]
ResNet18~\cite{he2016deep} & 21.3M & 31.8 & 49.6 & 33.6 & 16.3 & 34.3 & 43.2\\
Pure-MLP-S & 17.6M &27.1 & 44.2 & 28.3 & 13.6 & 29.2 & 36.4 \\
ConvMLP-S & 18.7M & 37.2 & 56.4 & 39.8 & 20.1 & 40.7 & 50.4 \\
\midrule
ResNet50~\cite{he2016deep} & 37.7M & 36.3 & 55.3 & 38.6 & 19.3 & 40.0 & 48.8 \\
Pure-MLP-M & 25.9M &28.0 & 45.6 &29.0 &14.5 &29.9 &37.8 \\
ConvMLP-M & 27.1M & 39.4 & 58.7 & 42.0 & 21.5 & 43.2 & 52.5 \\
\midrule
ResNet101~\cite{he2016deep} & 56.7M & 38.5 & 57.8 & 41.2 & 21.4 & 42.6 & 51.1\\
Pure-MLP-L & 50.1M & 28.8 & 46.8 & 29.9 & 15.0 &31.0 &38.4 \\
ConvMLP-L & 52.9M & 40.2 & 59.3 & 43.3 & 23.5 & 43.8 & 53.3 \\
\bottomrule[2pt]
\end{tabular} 
\caption{Comparison between ConvMLP and ResNet as RetinaNet backbones on MS COCO.}
\label{tab:retinanet}
\end{table*}
\begin{table*}[!ht]
\centering
\begin{tabular}{l|c|ccc|ccc}
\toprule[2pt]
Mask R-CNN Backbone & \# Params & $AP^{b}$ & $AP^{b}_{50}$ & $AP^{b}_{75}$ & $AP^{m}$ & $AP^{m}_{50}$ & $AP^{m}_{75}$ \\ 
\midrule[1.5pt]
ResNet18~\cite{he2016deep} & 31.2M & 34.0 & 54.0 & 36.7 & 31.2 & 51.0 & 32.7 \\
Pure-MLP-S & 27.5M & 25.1 & 45.1 & 25.1 & 25.0  & 42.8  & 26.0 \\
ConvMLP-S & 28.7M & 38.4 & 59.8 & 41.8 & 35.7 & 56.7 & 38.2 \\
\midrule
ResNet50~\cite{he2016deep} & 44.2M & 38.0 & 58.6 & 41.4 & 34.4 & 55.1 & 36.7 \\
Pure-MLP-M & 35.8M  & 25.8 & 46.1 &25.8 & 25.6 & 43.5 & 26.5 \\
ConvMLP-M & 37.1M & 40.6 & 61.7 & 44.5 & 37.2 & 58.8 & 39.8 \\
\midrule
ResNet101~\cite{he2016deep} & 63.2M & 40.4 & 61.1 & 44.2 & 36.4 & 57.7 & 38.8 \\
Pure-MLP-L & 59.5M  & 26.5 & 45.0 &27.4 & 26.7 & 47.5 & 26.8 \\
ConvMLP-L & 62.2M & 41.7 & 62.8 & 45.5 & 38.2 & 59.9 & 41.1 \\
\bottomrule[2pt]
\end{tabular} 
\caption{Comparison between ConvMLP and ResNet as Mask R-CNN backbones on MS COCO.}
\label{tab:maskrcnn}
\end{table*}
\begin{table*}[!ht]
\centering
\begin{tabular}{l|c|c}
\toprule[2pt]
Semantic FPN Backbone & \# Params & mIoU \\ 
\midrule[1.5pt]
ResNet18~\cite{he2016deep} & 15.5M & 32.9 \\
Pure-MlP-S & 11.6M & 23.9 \\
ConvMLP-S & 12.8M & 35.8 \\
\midrule
ResNet50~\cite{he2016deep} & 28.5M & 36.7 \\
Pure-MlP-M & 19.9M & 25.2 \\
ConvMLP-M & 21.1M & 38.6  \\
\midrule
ResNet101~\cite{he2016deep} & 47.5M & 38.8 \\
Pure-MlP-L & 43.6M & 26.3 \\
ConvMLP-L & 46.3M & 40.0 \\
\bottomrule[2pt]
\end{tabular} 
\caption{Comparison between ConvMLP and ResNet as Semantic FPN backbones on ADE20k.}
\label{tab:semseg}
\end{table*}

\section{Salient Maps, Bias, and OverFitting}
In an effort to analyze the differences in ConvMLP we investigate the salient maps of MLP-Mixer, ResMLP, and ConvMLP. 
We created salient maps based on the final output. We then selected a random sample of images from the network and investigated what they looked like.
For the most part images had fairly similar maps, but in some images there were stark differences that we believe highlight major differences in the networks. 
While these images are hand selected we note that there are some counter examples.
Though the counter examples exist we found that there exists a trend within the biases, which we try to highlight here.
We show these selected images in Figure~\ref{fig:salient}. 
To the left we label each network. 
At the top of the column we provide the ground truth label for the image and underneath each image we provide the network's corresponding predicted label.
Specifically we show salient maps where the MLP-Mixer model contains 59.9M params, ResMLP contains 30M params, and we use ConvMLP-L with 42.7M params. 
Additionally, we note that the MLP-Mixer and ResMLP models are both versions that were trained from scratch on ImageNet, making for a more fair comparison. 
The first aspect that should be noted is that MLP-Mixer and ResMLP have more ``pixelated'' looking salient maps.
We can clearly see the effect of ResMLP's smaller patches and subsequently finer resolutions to where the network sees information.

Starting with the left most image, Binoculars, we notice some interesting things. 
For one, a human would likely not label this image as an example of binoculars.
We should note that several labels within the dataset exist within this image.
The second thing to notice is the salient maps. 
MLP-Mixer pays attention to almost everything except the car and predicts the correct label.
ResMLP highlights the hares and a little bit of the person (near the face) but predicts the incorrect label of car wheel.
Lastly, ConvMLP also pays attention to the hares but incorrectly classifies the images as hare.
This presents an interesting phenomena when analyzing results. 
We have a clue that MLP-Mixer and ResMLP might be overfitting the data.
While ConvMLP makes the misclassification, we note that it is at least paying attention to the part of the image that closely corresponds to the label that it predicts. 
MLP-Mixer and ResMLP offers more suspicious behavior as they pay attention to different parts of the image than what they predict.
This may suggest overfitting.

In the second image all three networks correctly identify the image as a ballpoint pen, but there are interesting things to note in the salient maps. 
MLP-Mixer and ConvMLP both ignore the rubix cube and we can see their salient maps outline the pens (possibly pencils). 
Between MLP-Mixer and ConvMLP we notice that ConvMLP pays slightly more attention to the pens, though MLP-Mixer pays more attention to the rest of the scene.
Additionally, we note that ResMLP pays more attention to the rubix cube that the other two networks. 

In the third image we see a worrying problem in MLP-Mixer. 
Ribeiro et al\cite{ribeiro2016why} showed how huskies can be misclassified as wolves due to the presence of snow within the image. 
That is, if an image contains snow then it would be far more likley to be classified as a wolf, showing the bias that the network learned. 
What is worrying here is that the areas MLP-Mixer pays significant attention don't contain dogs, such as the bottom center image and center left.
This actually gets worse with the larger Mixer model, not shown, and it exclusively pays attention to sub-images without dogs. 
This suggests that the network may be biased from the background in the image, which is likely given that images of malamutes are likely to have snow within the image.
Again, this suggests potential for overfitting.
ResMLP and ConvMLP do not make these same mistakes. 
Both ResMLP and ConvMLP pay more attention to the dogs in the right most column, which are more centered in their respective sub-pictures, but ConvMLP also pays attention to the dogs in the top row. 
Neither of these networks pay any significant attention to sub-images without dogs in them, which is a good sign.
Despite ResMLP and ConvMLP misclassifying the image, the labels they provide are acceptable, being very close to the true labels. 
Both Canadian Eskimo Dogs and Malamutes look very similar in appearance and humans commonly call both Huskies. 
We note here that ImageNet contains exclusively images of Canadian Eskimo Dogs and does not contain images of American Eskimo Dogs, which are more easily distinguishable. 

With the last column we see a similar story. 
None of the networks get the classification correct.
MLP-Mixer again pays significant attention to the majority of the image and specifically the background. 
While all three models pay attention to the top left background of the image it is clear that MLP-Mixer is focused less on the animals than the other two networks.
Again we note that Gazelles, Impalas, and Hartebeests have very similar appearances, and that all three networks predict reasonable labels.

We found many similar examples of these patterns while analyzing salient maps for these networks and saw these patterns worsen with the larger models, despite that they achieve significantly higher accuracy scores. 
We believe that this type of analysis suggests that these networks show significantly different biases.
We also believe that this analysis supports claims that MLP-Mixer is overfitting the dataset and that scale is not all one needs to perform well. 
Rather that scale harms performance, but not in the way that we are evaluating models.
With this analysis we believe that there is significant evidence that MLP-Mixer and ResMLP overfit the ImageNet dataset and that accuracy cannot be the only score used to evaluate a model's performance. 
With this analysis we encourage the reader to use Occam's razor when selecting models and encourage practitioners to perform similar and more inclusive analyses when evaluating models. 
This analysis highlights that just because one does well on accuracy does not mean that is has low bias or will generalize well to the real world. 
With this it becomes important to analyze models and understand the biases that they have as well as the biases within the dataset. 

\section{Object Detection \& Semantic Segmentation}
We provide details of results on MS COCO and ADE20K benchmarks as reference to Figure~\ref{fig:downstream}. For MS COCO, we use ResNet, Pure-MLP and ConvMLP as backbone of RetinaNet and Mask R-CNN. The results are shown in Table~\ref{tab:retinanet} and Table~\ref{tab:maskrcnn}. For ADE20k, we use ResNet, Pure-MLP and ConvMLP as backbone of Semantic FPN and the result is shown in Table~\ref{tab:semseg}.

\end{document}